\documentclass{article}
\newcommand{\undertitle}[1]{}

\usepackage{arxiv}

\usepackage[utf8]{inputenc}
\usepackage[T1]{fontenc}
\usepackage[hidelinks]{hyperref}
\usepackage{url}
\usepackage{booktabs}
\usepackage{amsmath}
\usepackage{amsfonts}
\usepackage{microtype}
\usepackage{graphicx}
\usepackage{float}
\usepackage[numbers]{natbib}
\usepackage{doi}
\usepackage{array}
\usepackage{tabularx}
\usepackage{xcolor}
\usepackage{fancyhdr}
\usepackage{wrapfig}
\newcommand{\headeright}{}

\newcolumntype{Y}{>{\raggedright\arraybackslash}X}

\title{Technical Report on the Turba Fertilizer Machine Learning Stack in Morocco}

\author{%
  \parbox{0.96\textwidth}{%
    \centering 
    \small 
    \href{https://orcid.org/0000-0002-1581-3958}{\includegraphics[scale=0.065]{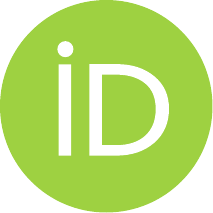}\hspace{1mm}Abdelghani Belgaid}
    \quad
    \href{https://orcid.org/0009-0004-6015-325X}{\includegraphics[scale=0.065]{orcid.pdf}\hspace{1mm}Zakaria Mahmoud}
    \quad
    \href{https://orcid.org/0009-0002-9091-9645}{\includegraphics[scale=0.065]{orcid.pdf}\hspace{1mm}Fahd Chibani}
    \\[1.5mm]
    \href{https://orcid.org/0009-0009-8216-2962}{\includegraphics[scale=0.065]{orcid.pdf}\hspace{1mm}Oumnia Ennaji}
    \quad
    \href{https://orcid.org/0009-0009-9930-7419}{\includegraphics[scale=0.065]{orcid.pdf}\hspace{1mm}Younes Boudoul}
    \quad
    \href{https://orcid.org/0009-0006-9814-4476}{\includegraphics[scale=0.065]{orcid.pdf}\hspace{1mm}Dounia Rachid}
    \\[4mm]
    \normalsize Turba Open Lab
  }%
}

\date{}

\begin{document}
\maketitle

\begin{abstract}
Site-specific fertilizer recommendation systems adapt nutrient advice to location, soil properties, crop type, and production targets, but scientific reuse is constrained when recommendation functions remain accessible mainly through interactive interfaces, outputs are not versioned, and trained approximations cannot be independently loaded or benchmarked. This technical report presents the Turba fertilizer machine learning stack, a three-layer open-source implementation for reproducible site-specific fertilizer recommendation in Morocco. \texttt{turba-client} provides programmatic access to publicly accessible site profiles, crop-specific target-yield spaces, and N, P$_2$O$_5$, and K$_2$O recommendation workflows; \texttt{turba-data} distributes analysis-ready snapshots; and \texttt{turba-models} packages crop-specific machine learning surrogates of recommendation outputs. The architecture links upstream retrieval, versioned analytical snapshots, reproducible cross-model benchmarking, and loadable offline surrogates while preserving the distinction between recommendation-system outputs, observed agricultural data, and model-generated predictions. The first dataset was constructed from 44,096 unique ESA WorldCereal locations. Scenario expansion across supported cereal workflows generated 132,017 crop-location recommendation requests under a medium target-yield setting. The resulting 22-variable dataset spans 10 regions, 66 provinces, and 1,149 communes. Nine regression families were evaluated under a fixed deterministic 80/20 protocol, and the current release packages five best-performing crop-specific models. The machine learning task is recommendation-function emulation rather than prediction of observed crop response. The stack provides a reproducible basis for spatial and temporal validation, uncertainty estimation, field-trial comparison, and future integration with additional data.
\end{abstract}


\section{Introduction}

Site-specific nutrient management requires fertilizer decisions to account for heterogeneity in soil fertility, crop requirements, production targets, and environmental conditions. Machine learning has become increasingly relevant for the management of nutrients because nonlinear relationships across soil, crop, spatial, and management variables can be represented computationally and incorporated into decision-support workflows \citep{ennaji2023machine}. Previous Moroccan research has developed machine learning and optimization methods for site-specific fertilizer recommendation \citep{abdelghani2025site,ennaji2026machine} and has demonstrated substantial spatial and temporal heterogeneity in cereal production systems \citep{ennaji2025soilvariables}.

The availability of computational methods alone is insufficient for reproducible fertilizer research. Recommendation systems accessed through interactive interfaces are difficult to incorporate into automated experiments, while analyses reconstructed repeatedly from mutable upstream services may not yield identical research objects. Reproducible computational science therefore requires preservation of data provenance, executable procedures, software versions, and intermediate artifacts \citep{sandve2013ten}. The FAIR principles similarly emphasize accessibility, interoperability, reusability, persistent identification, and explicit provenance for data, algorithms, and workflows \citep{wilkinson2016fair}. Robust research software extends these requirements to testing, dependency management, documentation, and recoverable execution environments \citep{taschuk2017ten}.

These considerations are particularly relevant in Morocco. Fertimap is Morocco's soil-fertility mapping and fertilization decision-support system, developed by OCP Group in collaboration with a consortium of agronomic research institutions and the Moroccan Ministry of Agriculture \citep{fertimap2026}. A national digital soil-mapping study reported that earlier Fertimap soil-property maps were viewable through a public interface but that the underlying raster products were not downloadable for independent reuse, illustrating the distinction between public visualization and reusable computational access \citep{bouslihim2025baseline}. Open Earth-observation resources such as ESA WorldCereal can complement such systems by providing reproducible spatial crop information for the construction of agricultural datasets \citep{van2023worldcereal}.

The Turba fertilizer machine learning stack is an independent open-source initiative that builds reproducible computational resources around publicly accessible Fertimap recommendation workflows through three components: \texttt{turba-client} provides programmatic access, \texttt{turba-data} distributes versioned analysis-ready recommendation datasets, and \texttt{turba-models} provides pretrained machine-learning surrogates and benchmark artifacts \citep{turba_open_lab_2026_20044002}. Its central design principle is provenance separation. Values retrieved from Fertimap, values stored in a versioned dataset, and values generated by a learned surrogate are treated as distinct scientific objects. Turba is independently developed and is not affiliated with or endorsed by Fertimap, OCP Group, or the Moroccan Ministry of Agriculture. The stack is intended for reproducible analysis and methodological research; no component is presented as independent evidence of fertilizer efficacy or field-level agronomic optimality.

\section{System Architecture and Computational Provenance}

Let
\begin{equation}
\mathbf{x}=
(Long,Lat,\mathrm{pH},\mathrm{OM},P_{\mathrm{available}},K_{\mathrm{available}})
\end{equation}
denote the geographic and soil state used by the current machine learning layer, where $Long$ and $Lat$ are longitude and latitude, $\mathrm{OM}$ is soil organic matter, and $P_{\mathrm{available}}$ and $K_{\mathrm{available}}$ denote available phosphorus and potassium. Let $c$ denote a crop workflow and $t$ a target-yield specification. The upstream recommendation process is represented abstractly as
\begin{equation}
\mathbf{y}=g(\mathbf{x},c,t),
\qquad
\mathbf{y}=(N,P_{2}O_{5},K_{2}O),
\label{eq:upstream}
\end{equation}
where $g(\cdot)$ denotes the externally maintained recommendation function.

\begin{figure}[h]
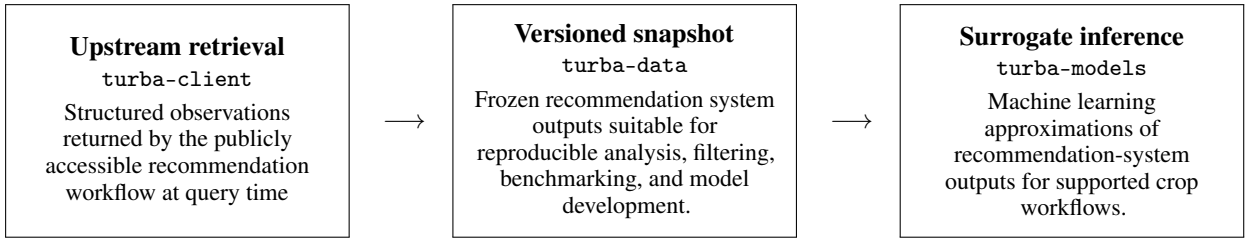

\centering
\setlength{\fboxsep}{8pt}
\fbox{\parbox[c][2.5cm][c]{0.24\textwidth}{\centering\textbf{Upstream retrieval}\\[0.5mm]\small\texttt{turba-client} \\ [1mm] Structured observations returned by the publicly accessible recommendation workflow at query time}}
\quad$\longrightarrow$\quad
\fbox{\parbox[c][2.5cm][c]{0.24\textwidth}{\centering\textbf{Versioned snapshot}\\[0.5mm]\small\texttt{turba-data} \\ [1mm] Frozen recommendation system outputs suitable for reproducible analysis, filtering, benchmarking, and model development.}}
\quad$\longrightarrow$\quad
\fbox{\parbox[c][2.5cm][c]{0.24\textwidth}{\centering\textbf{Surrogate inference}\\[0.5mm]\small\texttt{turba-models} \\ [1mm] Machine learning approximations of recommendation-system outputs for supported crop workflows.}}
\caption{Computational architecture of the Turba fertilizer machine learning stack. Each layer produces a provenance-distinct research object: an upstream response at query time, a frozen analytical record, or a machine learning prediction.}
\label{fig:architecture}
\end{figure}

The stack separates three operations around this function (Figure~\ref{fig:architecture}). \texttt{turba-client} evaluates the publicly accessible upstream workflow and returns structured site and recommendation information. \texttt{turba-data} stores selected evaluations of $g(\cdot)$ as versioned Parquet snapshots,
\begin{equation}
\mathcal{D}=\{(\mathbf{x}_i,c_i,t_i,\mathbf{y}_i)\}_{i=1}^{n},
\label{eq:dataset}
\end{equation}
thereby decoupling downstream analysis from subsequent changes in upstream availability or implementation. \texttt{turba-models} distributes learned functions
\begin{equation}
\widehat{\mathbf{y}}=f_{\theta,c}(\mathbf{x})
\label{eq:surrogate}
\end{equation}
that approximate recommendation outputs for supported crop workflows. The learned function $f_{\theta,c}$ is therefore a surrogate of $g$, not a crop-response model and not an independently validated agronomic recommendation system.

\subsection{Data-access layer: \texttt{turba-client}}

\texttt{turba-client} supports Python~$\geq$3.9 and exposes the \texttt{TurbaClient} interface together with a command-line interface. Geographic coordinates can be used to retrieve administrative and soil context through \texttt{get\_site\_profile()}, discover valid crops and associated target-yield ranges through \texttt{list\_crops()}, and retrieve fertilizer recommendations through \texttt{get\_recommendations()}. Batch execution is supported through \texttt{get\_recommendations\_batch()}, while \texttt{prepare\_input\_table()} normalizes heterogeneous tabular schemas. Coordinate bounds, target-yield levels, crop availability, required columns, and crop-specific target-yield ranges are validated before execution. Explicit exceptions distinguish validation failures, unavailable crops, missing site information, and upstream-response failures \citep{turba_open_lab_2026_20044002}.

The client accepts controlled overrides of soil pH, organic matter, available P$_2$O$_5$, and available K$_2$O. These overrides permit sensitivity analysis and comparison between upstream values and independently measured or curated inputs. Because this layer depends on an external service, exact reproduction of future requests cannot be guaranteed after upstream changes.

\subsection{Data layer: \texttt{turba-data}}

\texttt{turba-data} supports Python~$\geq$3.10 and exposes \texttt{list\_datasets()} and \texttt{load\_dataset()}. Dataset artifacts are distributed as Parquet files with registry metadata. Packaging each snapshot separately from the data-access layer establishes a stable analytical object and permits experiments to be reproduced without repeating upstream requests. The first registered dataset is \texttt{esa\_worldcereal\_morocco\_cereals\_medium} \citep{turba_open_lab_2026_20044002}.

\subsection{Model layer: \texttt{turba-models}}

\texttt{turba-models} supports Python~$\geq$3.10 and provides \texttt{list\_models()}, \texttt{load\_model()}, \texttt{predict\_\allowbreak recommendation()}, and \texttt{regression\_\allowbreak report()}. Serialized model artifacts are distributed with a registry defining crop workflow, estimator family, input schema, output schema, split policy, and evaluation metadata. The current feature vector contains longitude, latitude, soil pH, organic-matter percentage, available P$_2$O$_5$, and available K$_2$O. The three model outputs are recommended N, P$_2$O$_5$, and K$_2$O \citep{turba_open_lab_2026_20044002}. The six predictors constitute only the model input schema; the 22-variable packaged dataset additionally retains recommendation outputs and contextual, administrative, crop-scenario, target-yield, and provenance fields used for analysis and traceability.

\section{Dataset Construction and Coverage}

Agricultural locations were derived from ESA WorldCereal, an open global crop and irrigation mapping system designed for seasonal and reproducible Earth-observation analysis \citep{van2023worldcereal}. The retained Moroccan subset contained 43,825 unique locations labelled \texttt{Winter Cereals} and 271 unique locations labelled \texttt{Maize}, yielding 44,096 unique geographic sites.

\begin{wraptable}{r}{0.45\textwidth}
\vspace{-10pt}
\centering
\caption{Construction and geographic scope of the first packaged dataset.}
\label{tab:data}
\footnotesize
\begin{tabularx}{\linewidth}{@{}Xr@{}}
\toprule
\textbf{Property} & \textbf{Value} \\
\midrule

\multicolumn{2}{@{}l}{\textbf{Source locations}} \\
Unique geographic sites & 44,096 \\
\quad Winter-cereal sites & 43,825 \\
\quad Maize sites & 271 \\

\addlinespace
\multicolumn{2}{@{}l}{\textbf{Recommendation scenarios}} \\
Total crop-location scenarios & 132,017 \\
\quad Winter-cereal scenarios & 131,475 \\
\quad Maize scenarios & 542 \\

\addlinespace
\multicolumn{2}{@{}l}{\textbf{Dataset coverage}} \\
Variables & 22 \\
Regions & 10 \\
Provinces & 66 \\
Communes & 1,149 \\
Target-yield setting & Medium \\

\bottomrule
\vspace{-4pt}
\end{tabularx}
\end{wraptable}

WorldCereal classes are broader than the crop workflows represented in the upstream recommendation system. The source labels were therefore used only to define a set of compatible recommendation scenarios rather than to infer exact field-level crop identities. The three winter-cereal workflows, Wheat (Rainfed), Wheat (Irrigated), and Barley (Rainfed), were selected because they are the supported cereal workflows in the upstream system that are compatible with the broad Winter Cereals class. Likewise, Maize (Grain) and Maize (Silage) constitute the supported maize workflows. Enumerating all compatible workflows deliberately preserves uncertainty about the crop and irrigation regime rather than assigning an unsupported field-level identity. The resulting request universe was
\begin{equation}
43{,}825\times3+271\times2=132{,}017
\end{equation}
crop-location scenarios.

\begin{figure}[h]
\centering
\includegraphics[width=0.7\linewidth]{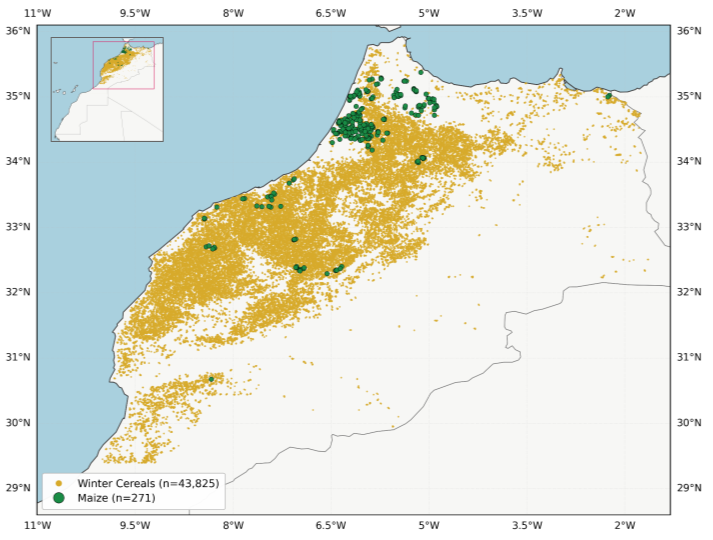}
\caption{Spatial distribution of the 44,096 unique source locations by ESA WorldCereal class: \texttt{Winter Cereals} (43,825) and \texttt{Maize} (271).}
\label{fig:spatial_coverage}
\end{figure}

A medium target-yield level was used for every scenario. Recommendation retrieval was implemented as a resumable process keyed by location and crop workflow. The packaged dataset contains 22 variables and covers 10 regions, 66 provinces, and 1,149 communes. Table~\ref{tab:data} summarizes its construction, while Figure~\ref{fig:spatial_coverage} shows the geographic distribution.

The sampling unit requires careful interpretation. The dataset contains 132,017 recommendation scenarios, not 132,017 independently observed agricultural fields. Three scenarios share each winter-cereal coordinate and two share each maize coordinate. Scenario expansion likewise does not establish that Wheat (Rainfed), Wheat (Irrigated), or Barley (Rainfed) was physically cultivated at each winter-cereal site. This distinction prevents artificial inflation of the effective spatial sample size and avoids interpreting analytical scenario labels as field observations.

The nominal spatial support is therefore 44,096 unique coordinates rather than 132,017 independent locations. Within the crop-specific modeling datasets, each winter-cereal workflow contains 43,825 unique coordinates and each maize workflow contains 271. Because models are fitted separately by crop workflow, a coordinate occurs only once within a given crop-specific modeling table, although the same coordinate can occur across different workflow-specific datasets. Consequently, statistical support and generalization should be interpreted in terms of unique sites and spatial dependence, not the expanded scenario count.

\section{Machine Learning Benchmark and Evaluation}

For the primary benchmark, each crop workflow was evaluated independently using a deterministic 80/20 partition of its crop-location records. Because each crop-specific table contains one record per coordinate, no identical coordinate is duplicated between training and test sets within a given crop workflow. The same geographic site can nevertheless occur in different workflow-specific datasets, and nearby sites remain spatially dependent; the benchmark should therefore be interpreted as an interpolation evaluation rather than evidence of geographic transfer.

Nine regression families were evaluated for direct emulation of fertilizer recommendation outputs: Extra Trees \citep{geurts2006extremely}, LightGBM \citep{ke2017lightgbm}, CatBoost \citep{prokhorenkova2018catboost}, Random Forest \citep{breiman2001random}, XGBoost \citep{chen2016xgboost}, Linear Regression, Ridge Regression \citep{hoerl1970ridge}, Elastic Net \citep{zou2005regularization}, and AdaBoost \citep{freund1997decision}. Models were trained using fixed library configurations without hyperparameter optimization; all estimator settings and random seeds are preserved in the released training workflow. Benchmark artifacts report performance by crop, model family, and nutrient target using $R^2$, mean absolute error (MAE), median absolute error, root mean squared error (RMSE), mean absolute percentage error (MAPE), and symmetric mean absolute percentage error (sMAPE); overall values are the unweighted means of the corresponding N, P$_2$O$_5$, and K$_2$O target-wise metrics \citep{turba_open_lab_2026_20044002}. 

The current release packages one surrogate for each of the five supported crop workflows: LightGBM for Barley (Rainfed), Wheat (Rainfed), and Wheat (Irrigated), and XGBoost for Maize (Grain) and Maize (Silage). Under the fixed deterministic 80/20 evaluation recorded in the model registry, the three cereal models achieved overall $R^2$ values above 0.9, with MAE ranging from 0.040 to 0.069 and RMSE from 0.160 to 0.210. The two maize models also showed high agreement with the reference recommendation outputs, with $R^2=0.987$ for Maize (Grain) and $R^2=0.994$ for Maize (Silage). Their corresponding MAEs were 0.385 and 0.478, respectively (Table~\ref{tab:models}).

\begin{table}[h]
\centering
\caption{Packaged machine-learning surrogates and overall evaluation results.}
\label{tab:models}
\small
\begin{tabular}{lrlrrr}
\toprule
\textbf{Crop} & \textbf{Sample} & \textbf{Model} &
\textbf{$R^2$} & \textbf{MAE (kg ha$^{-1}$)} & \textbf{RMSE (kg ha$^{-1}$)} \\
\midrule
Barley (Rainfed)  & 43,825 & LightGBM & 0.999 & 0.040 & 0.160 \\
Wheat (Rainfed)   & 43,825 & LightGBM & 0.999 & 0.058 & 0.188 \\
Wheat (Irrigated) & 43,825 & LightGBM & 0.999 & 0.069 & 0.210 \\
Maize (Grain)     & 271    & XGBoost  & 0.987 & 0.385 & 1.428 \\
Maize (Silage)    & 271    & XGBoost  & 0.994 & 0.478 & 1.453 \\
\bottomrule
\end{tabular}
\end{table}

These results indicate high-fidelity emulation of the reference recommendation function within the evaluated split, particularly for the much larger winter-cereal datasets. Nevertheless, they should be interpreted as surrogate fidelity, rather than agronomic predictive performance. Because the prediction task is $\widehat{\mathbf{y}}=f_{\theta,c}(\mathbf{x})\approx g(\mathbf{x},c,t)$, high $R^2$ and low absolute errors measure agreement with reference recommendation outputs; they do not measure crop yield response, nutrient-use efficiency, economic return, environmental loss, or agronomic optimality. Percentage-based metrics require additional caution. Fertilizer recommendation targets can contain zero or near-zero values, for which percentage errors become unstable or poorly interpretable \citep{hyndman2006another}. In the packaged dataset, exactly zero recommendations account for 0.00\%, 31.05\%, and 78.83\% of N, P$_2$O$_5$, and K$_2$O targets, respectively. Using 1~kg~ha$^{-1}$ as a near-zero threshold, the corresponding proportions are 0.00\%, 31.46\%, and 79.00\%. This concentration near zero helps explain why percentage-based metrics can remain large despite high $R^2$ and low absolute errors. This behavior is visible in the released benchmark artifacts, where MAPE can become extremely large despite simultaneous high $R^2$ and low absolute error. Although sMAPE is bounded and less sensitive to scale than MAPE, such targets remain difficult to summarize with percentage-based errors; $R^2$, MAE, RMSE, target distributions, and residual diagnostics are therefore emphasized here.

A second limitation concerns the evaluation design. Random row-wise partitioning is primarily an interpolation test. Geographic observations commonly exhibit spatial dependence, and conventional random cross-validation can underestimate prediction error when spatial or temporal structure is present \citep{roberts2017cross}. This concern is particularly relevant because longitude and latitude are explicit predictors and multiple crop scenarios share geographic coordinates. Spatial blocking should therefore use geographically separated site groups to reduce leakage arising from spatial dependence among nearby observations. Region-holdout evaluation and temporal validation are also required before geographic or temporal generalization can be claimed. The practical importance of this distinction has previously been demonstrated in Moroccan fertilizer modeling, where random-split yield prediction reached approximately $R^2=0.96$ while temporal evaluation declined to approximately $R^2=0.17$ \citep{ennaji2026machine}.

Finally, the maize subsets contain only 271 locations per workflow, compared with 43,825 for each winter-cereal workflow. The strong maize interpolation results should therefore not be interpreted as having the same statistical support as the winter-cereal results. Independent spatial validation, larger maize datasets, and comparison against fertilizer-response experiments remain necessary to establish robustness beyond the present recommendation-function emulation task. Accordingly, none of the benchmark metrics reported here constitute evidence that the reproduced fertilizer rates are agronomically optimal. They quantify only how closely an offline surrogate reproduces the outputs of the upstream recommendation function under the evaluated data distribution.

\section{Reproducibility and Availability}

Reproducibility is implemented at three levels. Software reproducibility is supported by versioned Python packages, automated tests, explicit dependencies, validated interfaces, and documented exceptions. Data reproducibility is supported by a registered Parquet snapshot that can be loaded without re-executing upstream requests. Model reproducibility is supported by serialized \texttt{.joblib} artifacts, a machine-readable model registry, benchmark reports, and the corresponding training workflow. The released benchmark workflow uses random seed 42 and fixed estimator configurations. The executable training notebook records the complete estimator settings and generates the target-level benchmark, overall leaderboard, selected-model summary, model registry, and serialized model artifacts from a single run. A version-pinned environment file records the Python and package dependencies required for reproduction. Starting from the registered \texttt{turba-data} snapshot, a user can reproduce the benchmark by installing the locked environment and executing \texttt{notebooks/esa\_worldcereal\_morocco\_cereals\_medium\_model\_training\_and\_evaluation.ipynb}. These design choices follow established recommendations for preserving computational provenance and maintaining robust research software \citep{sandve2013ten,taschuk2017ten}.

The source code is available through the \texttt{open-turba} organization at \url{https://github.com/open-turba/}. The three components are maintained at \url{https://github.com/open-turba/turba-client}, \url{https://github.com/open-turba/turba-data}, and \url{https://github.com/open-turba/turba-models}. Python distributions are published as \texttt{turba-client}, \texttt{turba-data}, and \texttt{turba-models}. Archived software releases are associated with Zenodo DOI \href{https://doi.org/10.5281/zenodo.20044002}{10.5281/zenodo.20044002} \citep{turba_open_lab_2026_20044002}. The released source repositories use the MIT License. This license applies to Turba software and does not transfer ownership, licensing rights, or redistribution rights over upstream services or third-party data sources.

\section{Limitations and Future Directions}

\subsection{Limitations}

The main technical limitation of \texttt{turba-client} is its dependency on the structure and availability of the upstream service. A versioned snapshot mitigates this dependency for completed experiments, but does not make future upstream requests immutable. The principal data limitation is the scenario-based crop expansion: broad WorldCereal classes support candidate workflow generation but do not constitute field-level confirmation of specific crop or irrigation status. The principal modeling limitation is the use of six static spatial and soil variables with a deterministic random split and no explicit uncertainty calibration.

\subsection{Near-term validation priorities}

Future validation should prioritize site-grouped spatial blocking, region-holdout evaluation, temporal transfer, uncertainty quantification, out-of-distribution detection, and reproducible benchmark protocols \citep{kallenberg2026cy,eda8992d0e5d476ca1dff61715835423}, alongside independent fertilizer-response experiments and prospective field trials evaluating yield response, nutrient-use efficiency, profitability, and environmental performance.

\subsection{Longer-term research roadmap}

The longer-term roadmap follows an \emph{observe $\rightarrow$ infer $\rightarrow$ decide $\rightarrow$ simulate $\rightarrow$ learn} architecture. \texttt{turba-data} will expand observations through additional fertilizer recommendation records from Moroccan parcels, crowdsourced field and soil records, and multimodal Earth-observation and environmental data, including Sentinel-1/2, MODIS, Landsat-8/9, global soil-moisture products, ERA5 and NASA data services, CropHarvest, and soil information from iSDAsoil, SoilGrids, and SoilHive \citep{ouassanouan2026crop,osm2026,van2023worldcereal,tseng2021cropharvest,hersbach2020era5,belgaid_2025_17699363,poggio2021soilgrids,hengl2021african,soilhive2026}. Field delineation and validation could incorporate Moroccan field-boundary labels derived from resources such as Fields of the World and progressively refined through user-contributed observations \citep{kerner2025fields}. A planned Turba UI would expose these resources through interactive mapping, scenario construction, uncertainty visualization, and structured community data collection, while \texttt{turba-client} continues to provide interoperability with external systems; extensions will also capture structured regional and generic-formula recommendations, including fertilizer products, application stages, rates, yield information, and costs, to support broader datasets and machine learning models \citep{turba_open_lab_2026_20044002}.

Biological soil information represents a complementary frontier. A proposed \texttt{turba-bio} layer could integrate soil microbiome taxonomic and functional information, including resources such as MGnify, to investigate whether microbial diversity and nutrient-cycling potential improve soil-state estimation and fertilizer decisions beyond conventional physicochemical measurements \citep{richardson2023mgnify}. Synthetic-data generation could complement empirical data collection in sparse-data settings, using general-purpose frameworks and agriculture-specific approaches to support augmentation, controlled experimentation, and robustness analysis \citep{patki2016synthetic,belgaid2025sagdaopensourcesyntheticagriculture}.

Finally, \texttt{turba-gym} is envisioned as a Gymnasium-compatible experimental environment for sequential nutrient-management and adaptive field experimentation, extending concepts demonstrated by CropGym \citep{towers2026gymnasium,overweg2021cropgym}. Coupling such environments with WOFOST/PCSE would enable reinforcement learning, active experimental design, climate and microclimate sensitivity analysis, and hybrid process--machine learning simulation \citep{van1989wofost,dewit_pcse}. Decision science analyses could further guide which information should be collected and how predictive systems should be deployed. Soil-measurement acquisition can, for example, be formulated as a value-of-information problem that balances measurement cost against expected downstream fertilizer-decision loss, while model-adoption concentration can be studied as a source of correlated prescription risk when many decision makers rely on similar algorithms \citep{belgaid2026monoculture}. Economic and food-system variables, following approaches such as NASA Harvest's Harvest2Market, could subsequently extend optimization from field-level nutrient prescriptions toward multi-objective decisions incorporating production, input costs, environmental risk, and market conditions \citep{harvest2market2026}. Across these extensions, observed, crowdsourced, synthetic, simulated, upstream, and model-generated quantities should remain provenance-distinct. 

\section{Conclusion}

The Turba fertilizer machine learning stack formalizes three distinct computational objects for site-specific fertilizer recommendation research in Morocco: recommendation responses, versioned analytical snapshots, and offline machine learning surrogates. The first dataset contains 132,017 crop-location scenarios derived from 44,096 unique WorldCereal locations, and the first model release provides five crop-specific surrogates alongside benchmarks spanning nine regression families. The principal contribution is the preservation of the computational provenance across these layers. Recommendation-system outputs are not presented as observed field responses, scenario-expanded crop labels are not presented as ground-truth crop identities, and surrogate agreement is not presented as agronomic validation. This separation provides a reproducible basis for benchmarking and method development while defining the validation requirements necessary for spatial transfer, temporal generalization, uncertainty-aware inference, crop simulation, and prospective fertilizer field trials.

\section*{Acknowledgments}

The Fertimap website and the recommendation workflows accessed by \texttt{turba-client} are independently maintained upstream resources. No ownership of the Fertimap service, its underlying data, or its recommendation methodology is asserted by the Turba fertilizer machine learning stack. ESA WorldCereal is acknowledged as the source of the crop-location information used in constructing the first packaged dataset.

\bibliographystyle{unsrtnat}
\bibliography{references}

\end{document}